# Building Energy Efficiency through Advanced Regression Models and Metaheuristic Techniques for Sustainable Management

Hamed Khosravi [1,*], Hadi Sahebi [2], Rahim Khanizad [3], Imtiaz Ahmed [4]

[1]*School of Industrial Engineering, Iran University of Science and Technology, Tehran, Iran, hamedkhosravi181@gmail.com*

[2]*School of Industrial Engineering, Iran University of Science and Technology, Tehran, Iran, hadi_sahebi@iust.ac.ir*

[3]*School of Industrial Engineering, Iran University of Science and Technology, Tehran, Iran, Khanizad@iust.ac.ir*

[4]*Department of Industrial & Management Systems Engineering, West Virginia University, Morgantown, WV 26505, imtiaz.ahmed@mail.wvu.edu*

*\*Corresponding Author: hamedkhosravi181@gmail.com*

**Abstract**

In the context of global sustainability, buildings are significant consumers of energy, emphasizing the necessity for innovative strategies to enhance efficiency and reduce environmental impact. This research leverages extensive raw data from building infrastructures to uncover energy consumption patterns and devise strategies for optimizing resource use. We investigate the factors influencing energy efficiency and cost reduction in buildings, utilizing Lasso Regression, Decision Tree, and Random Forest models for accurate energy use forecasting. Our study delves into the factors affecting energy utilization, focusing on primary fuel and electrical energy, and discusses the potential for substantial cost savings and environmental benefits. Significantly, we apply metaheuristic techniques to enhance the Decision Tree algorithm, resulting in improved predictive precision. This enables a more nuanced understanding of the characteristics of buildings with high and low energy efficiency potential. Our findings offer practical insights for reducing energy consumption and operational costs, contributing to the broader goals of sustainable development and cleaner production. By identifying key drivers of energy use in buildings, this study provides a valuable framework for policymakers and industry stakeholders to implement cleaner and more sustainable energy practices.

## 1. Introduction

Over the last few decades, there has been a marked increase in the use of energy in developing countries, a trend that is expected to continue in the near future [1]. The importance of energy conservation has recently received significant attention from a number of stakeholders, including governments, industry, academia, and other organizations. It is believed that this growing interest in energy conservation is a consequence of the growing demand for energy and the declining supply of energy resources [2]. For instance, in China, building energy consumption accounted for 28% of the total energy consumption in 2011 [3], and accounted for 21% of total primary energy consumption in 2020, with 1.06 billion tce consumed [4]. Similarly, in the United States, building energy consumption comprises approximately 39% of the total energy consumption [5]. According to the International Energy Agency (IEA), residential and commercial buildings are responsible for 32% of final energy consumption [6]. Many buildings consume approximately 20% more energy than is required as a result of incomplete construction or a failure to follow the intended design. The issue arises when facilities are not operated as originally designed, resulting in inefficient energy use [7].

In recent years, people have been spending more and more time indoors [8,9], and residential energy consumption has increased significantly. Buildings contribute significantly to global energy consumption and greenhouse gas emissions. To mitigate these negative impacts, buildings must adopt energy-efficient and sustainable practices [10]. So, predicting building energy consumption is essential for building managers to make better decisions that improve energy utilization rates [11]. However, predicting building consumption is difficult because climate, population, and seasonal variations create nonlinear patterns [12]. Currently, buildings provide data that can be used to extract useful insights, patterns, or knowledge from them [13]. It also provides an opportunity to uncover hidden data, improve our understanding of energy usage, and create strategies for reducing energy consumption. Nevertheless, extracting valuable information from the data can be challenging without advanced data analysis techniques [14-16].

Developing prediction models often involves the use of popular methods such as machine learning and artificial intelligence-based approaches [17-19]. Models based on machine learning are more effective at capturing the complex relationships between building-level characteristics and energy consumption since they have fewer restrictions regarding the statistical relationships among variables [20]. The algorithms employed for developing energy consumption prediction models possess certain advantages and disadvantages [21]. The commonly used supervised machine learning algorithms for model training include SVM, ANN, decision trees, and other statistical algorithms [22]. In addition to its flexibility, decision tree algorithm can be improved as the amount of training data increases [23]. The use of data-driven models presents a practical approach to predict energy consumption [24].

Research in this field has primarily focused on developing models or identifying factors that influence energy consumption. There have, however, been a limited number of studies that have attempted to simultaneously address both of these objectives. Furthermore, most of these studies have relied on a specific type of features and relatively smaller datasets, resulting in a lack of adequate representation of households in the data. To fill this knowledge gap, this paper makes its primary contribution by undertaking the following aspects:

- This study fills a significant gap in the existing literature by using a new and previously unused dataset. To the best of our knowledge, this dataset has never been used in a published study before, thus making this study unique in its examination of a large number of households and a wide range of effects. The extensive dataset captures a comprehensive representation of the target population, which leads to more robust and reliable findings.
- Secondly, this study considers various algorithms, including Lasso, Decision Tree, and Random Forest. This leads to better understand the data and build more robust models. Different feature selection techniques to optimize the combination of features and models have been used, as well. This contribution lies in the comparison and evaluation of these models in the specific context of the research questions.
- In addition, unlike most studies in this area, this study considers a variety of features, including financial aspects, utility information features, performance indicator features, building characteristics, and customer information features. As a result of assessing a variety of feature types, we contribute to the development of a more inclusive and adaptable framework for analyzing energy consumption.
- Finally, the scope of this research goes beyond the majority of the literature focusing on a single target variable. Using multi-dimensional analysis, potential strategies for sustainability and resource optimization by identifying factors that influence energy consumption and associated costs can be analyzed.

In short, this study utilizes three different models (lasso regression, decision tree, and random forest) to predict energy consumption and cost in a real-world dataset. More specifically, this study aims to identify and rank the factors that affect energy consumption and related costs in buildings. Data visualization is used to observe and uncover valuable relationships. Three different strategies are used to select features for the study. After predicting with the selected models and features, different scenarios are evaluated and compared with several criteria. Using genetic algorithm, the performance of the decision tree algorithms for prediction are improved. Finally, an analysis of energy consumption and cost reduction based on building variables is discussed considering the most efficient models identified following the evaluation.

## 2. Literature Review

In recent times, there has been significant focus on predicting the energy consumption of buildings [25], which has led to the development and implementation of various approaches for tackling related problems [26]. To improve the operational performance of building energy systems, data mining techniques are commonly employed to extract meaningful information from large sets of building operation data [27]. Typically, these methods fall into two main categories: supervised and unsupervised [28]. The use of data mining technologies in the building industry has been extensively studied over the last ten years, with several literature reviews published on the subject.

A decision tree algorithm was utilized by Yang G et al. in 2010 to optimize building energy consumption [29]. In the same year, N. Giatani et al. analyzed energy consumption data from 1100 schools to identify patterns at the building level. They used clustering techniques (k-means) and Matlab software to define heating energy consumption information in five clusters, which were analyzed [30]. In 2011, Wall et al. used hierarchical clustering algorithms to diagnose faults in HVAC systems, aiming to identify operational patterns [31]. The same year, R.S. Jota and colleagues used hierarchical clustering to predict building electricity consumption by identifying common consumption patterns [32]. F.W. Yu and colleagues used clustering techniques and SPSS software to evaluate the behavior and performance of the chiller system in two studies conducted in 2012 [33]. In the same year, a decision tree algorithm was applied to predict peak electrical energy demands [34]. To predict lighting energy consumption in 2013, Liu D and colleagues compared artificial neural networks with SVMs [35]. In the same year, Kavousian et al. analyzed the electricity consumption of 1,628 households and found that weather and the physical characteristics of the building had a greater impact on consumption than occupant behavior [36].

A year after, Tang et al. used the k-means algorithm to cluster the entire data before developing prediction models. They claimed that this approach reduces the prediction error and the computational burden. They employed this approach for modeling and predicting HVAC systems [37]. In 2015, hierarchical clustering was utilized to identify typical energy consumption patterns, demonstrating that the proposed method can effectively predict energy consumption and peak demand with high accuracy [38]. A method based on artificial neural networks was proposed by Deb et al. (2016) for forecasting cooling load in the building sector in the presence of data related to energy consumption. The authors used $R^2$ value to evaluate the results [39]. In the same year, Huebner et al. examined factors influencing electricity consumption in gas-heated residential buildings using data from 845 English households. Appliance ownership and usage, along with household size, were found to be the most influential variables [40].

Li et al. in 2017 compared four machine learning models in order to forecast the energy consumption in a retail building. As a result of their analysis, the Extreme Learning Machine (ELM) model was found to be the most efficient model in terms of forecasting [41].

In 2018, Ma et al. evaluated building energy performance. They utilized a hierarchical clustering algorithm in their research. In addition, they used advanced techniques such as dendrograms and heat maps to understand energy consumption behaviors in the building [42]. Liu et al. conducted accuracy analyses and compared models in 2019. The accuracy analyses were based on different types of buildings. As epidemic models, they compared artificial ANNs and SVMs based on their prediction process complexity, the accuracy of the results, and the number of inputs required [43]. One of the crucial aspects of machine learning models is parameter tuning. Consequently, Seyedzadeh et al. proposed a method for optimizing machine learning models for predicting heating and cooling loads in building energy consumption. This method employed multi-objective optimization techniques with evolutionary algorithms to explore the parameter space [44]. Random Forest, a widely used and significant machine learning model, was utilized by Pham et al. in the same year for short-term energy consumption prediction. The proposed model estimated multiple buildings' hourly energy consumption [45].

In 2021, by combining ensemble learning and pattern categorization, Dong et al. were able to predict an office building's hourly energy consumption [46]. A prediction model based on machine learning in the same year was trained using a vast dataset consisting of 3-month hourly data for 5760 energy-use cases that encompass various combinations of building characteristics, outdoor weather conditions, and occupant behaviors. Four machine learning algorithms were evaluated and compared during the model development process based on their prediction accuracy and computational efficiency [47]. Based on the analysis of architectural characteristics based on data mining, Shan et al. in 2022 identified the critical attributes of various types of buildings. In their study, Principal Component Analysis (PCA) and Random Forest Analysis were used to identify significant architectural characteristics associated with various levels of energy consumption [48]. The study conducted by Li et al. was a case analysis of an educational building. They introduced a novel method for forecasting building electricity load that involves using similarity judgement and an improved TrAdaBoost algorithm (iTrAdaBoos) and found that their proposed method had a simple structure, making it easy to implement for engineering purposes, compared to other advanced models [49]. In 2023, a supervised machine-learning model was developed by kapp et al. using data from 45 manufacturing plants, which were obtained from industrial energy audits. The goal was to create a general predictor of industrial building energy consumption [50]. In another research, an energy consumption benchmark for university buildings in Brazil was established. Three machine learning techniques were evaluated for this purpose, and SVM method was found to have the lowest mean absolute error and root mean absolute error. As a result, the SVM method was chosen to develop the benchmark and efficiency scales [51]. The summary of the recent literature can be seen in Table 1.

Table 1. A list of data mining-based methods for predicting building energy loads.

| Ref. | Year | Algorithm | Focus |
|---|---|---|---|
| [29] | 2010 | DT | Optimizing building energy consumption |
| [30] | 2010 | K-means | Identifying the patterns of energy consumption |
| [31] | 2011 | Hierarchical Clustering | Identify the operational patterns |
| [32] | 2011 | Hierarchical Clustering | Predicting building electricity consumption |
| [33] | 2012 | Clustering techniques | Examining the chiller system |
| [34] | 2012 | DT | Predicting peak electrical energy demands |
| [35] | 2013 | SVM, ANN | Prediction of lighting energy consumption |
| [36] | 2013 | Weighted regression model | Investigating the structural and behavioral factors that influence residential electricity consumption |
| [37] | 2014 | K-means | Modeling and predicting HVAC |
| [38] | 2015 | Hierarchical Clustering | Predicting energy consumption and peak demand |
| [39] | 2016 | ANN | Forecasting cooling load |
| [40] | 2016 | Regression models | Analyzing the extent to which building characteristics explain annual electricity consumption |
| [41] | 2017 | Backward propagation neural network (BPNN), support vector regression (SVR), adaptive network-based fuzzy inference system (ANFIS) and ELM | Forecasting the energy consumption in a retail building |
| [42] | 2018 | Hierarchical Clustering | Understanding energy consumption behaviors |
| [43] | 2019 | ANN, SVM | Accuracy analyses and model comparisons |
| [44] | 2020 | Multi-objective optimization techniques with evolutionary algorithms | Predicting heating and cooling loads |
| [45] | 2020 | RF | Short-term energy consumption prediction |
| [46] | 2021 | Ensemble learning models | Predicting an office building's hourly energy consumption |
| [47] | 2021 | Classification and regression trees (CART), ensemble bagging trees (EBT), ANN, and deep neural networks (DNN) | Occupant-behavior-sensitive energy consumption prediction |
| [48] | 2022 | PCA, RF | Identifying the significant architectural characteristics |
| [49] | 2022 | iTrAdaBoos | Forecasting building electricity load |

| Ref. | Year | Algorithm | Focus |
|------|------|-----------|-------|
| [50] | 2023 | A supervised machine-learning model (SVM) | Creating a general predictor of industrial building energy consumption |
| [51] | 2023 | Multiple linear regression (MLR), SVM, and ANN | Establishing an energy consumption benchmark |

## 3. Methodology

The current study employed the CRISP-DM methodology, which is an industry-agnostic process model commonly used in data mining. It consists of six iterative phases that direct the data mining process, beginning with a comprehension of the business context and concluding with the implementation of the outcomes [52-55]. By providing a structured approach to data mining, the CRISP-DM methodology can reduce the cost and time associated with data mining projects. Additionally, the methodology minimizes the knowledge requirements for data mining projects by creating a framework that can be used by individuals of varying levels of expertise [56]. Hence, it has been adapted in this study. Each phase is thoroughly described in this section, along with its utilization in the current study.

### 3.1. Business Understanding

One of the crucial stages in a data mining project is comprehending the business context. It dictates the data to be gathered, the analysis techniques to be employed, and the manner in which the findings should be presented [52]. The findings of this study have several implications for businesses and policymakers. The study examines a number of factors that can affect energy consumption, including building characteristics, occupancy patterns, and financial aspects. This information can be used by businesses to identify buildings where they can improve energy efficiency.

### 3.2. Data Understanding

During this phase, data is gathered, explored, and described while ensuring its quality. The task of describing the data can involve the use of statistical analysis techniques to identify attributes and correlations, as specified in the user guide [52]. For this study, the data was obtained from New York State Office of Information Technology Services [57], which includes 26 columns and 57925 rows pertaining to a particular building plan. As part of the Existing Residential Home Design initiative, Goldstar building performance contractors were hired to implement and construct comprehensive energy efficiency enhancements. Table 2 provides a description of the database.

**Table 2: Description of the dataset**

| Variable name | Data type | Variable description |
|---|---|---|
| Reporting Period | Numeric(ordinal) | Date reported |
| Home Performance Project ID | categorical | Unique project ID |
| Home Performance Site ID | categorical | Unique house location ID |
| Project County | categorical | The area where the project was done |
| Project City | categorical | The city where the project was done |
| Project Zip | categorical | Project zip code |
| Gas Utility | categorical | Name of gas supplier for the project site |
| Electric Utility | categorical | Name of electricity supplier for the project site |
| Project Completion Date | Numeric(ordinal) | Project completion date |
| Customer Type | categorical | Incentives or subsidies paid by the government |
| Low-Rise or Home Performance Indicator | categorical | The type of building in the project (each has benefits in terms of receiving facilities and loans) |
| Total Project Cost | Numeric(integer) | Project cost in US dollars |
| Total Incentives | Numeric(integer) | Financial incentives received by the building owner |
| Type of Program Financing | categorical | Indicates the type of program financing (if it is empty, ie it does not support conventional programs) |
| Amount Financed Through Program | Numeric(integer) | Project loan amount in dollars |
| Pre-Retrofit Home Heating Fuel Type | categorical | The type of fuel used in the heating system before the building was remodeled. |
| Year Home Built | Numeric(ordinal) | Date of construction of the house |
| Size of Home | Numeric(integer) | House area in square feet |
| Volume of Home | Numeric(integer) | Approximate volume of home air conditioning |
| Number of Units | Numeric(integer) | Approximate volume of home air conditioning |
| Measure Type | categorical | The main improvement of the project is on which part of the building |
| Estimated Annual kWh Savings | Numeric(integer) | Annual electrical storage in kilowatt hours |
| Estimated Annual MMBtu Savings | Numeric(integer) | Annual primary fuel storage in MMBtu |
| First Year Energy Savings $ Estimate | Numeric(integer) | Estimate the amount of cost saved in dollars |
| Homeowner Received Green Jobs-Green NY Free/Reduced Cost Audit (Y/N) | categorical | Indicates whether the landlord has used the plan (Green Jobs-Green NY Free / Reduced Cost Audit) |

### 3.3. Data Preparation

The process of preparing data for analysis involves employing data mining techniques. This phase typically takes up a significant amount of time during the analysis. It encompasses activities such as merging, cleansing, converting, and downsizing data [52]. The present study outlines three crucial steps within this phase: data transformation, data correction, and data reduction.

#### 3.3.1. Data transformation

In order to improve the data presentation in the project management system, various changes have been implemented to specific columns. For instance, the Project Completion Date column now only shows the year of completion, omitting the month and day information. Additionally, the Customer Type column has been modified to use the numbers 1 and 0 to indicate "assisted" and "market" respectively. Similarly, the Low-Rise or Home Performance Indicator column now uses the numbers 1 and 0 to denote "Home Performance Indicator" and "Low-Rise" correspondingly. Previously, the Homeowner Received Green Jobs-Green NY Free / Reduced Cost Audit (Y / N) column had only two categories, but it now includes the numbers 1 and 0 to signify "usage" and "non-usage" respectively. To simplify the data analysis, categorical columns with more than two categories have been transformed into more intelligible columns by incorporating dummy variables into the software.

#### 3.3.2. Data correction

During our analysis, we observed inconsistencies in the capitalization of the word "Gas" in the phrase "Natural Gas" in some instances. As the software is sensitive to such differences and treats them as separate lines, we standardized the capitalization by replacing instances of lowercase "gas" with uppercase "Gas". Additionally, we noticed a row with the value 1347 in the electric type column. As there was no discernible difference between the two categories, we assumed this value also belonged to the same category. Therefore, we treated it accordingly. Furthermore, as per the description provided, the Type of Program Financing column was identified to have empty values indicating that conventional financial resources were not supported. Consequently, we filled these empty values with "not financed".

#### 3.3.3. Data reduction

As a first step, the Location column was removed from the dataset as it contained no useful information for the problem. The columns displaying data submission date, project ID, home location ID, and project codec were also excluded, as they were deemed irrelevant. It was decided to create only ten new columns using dummy variables to address the issue of categorical columns with many categories. This approach was adopted to avoid excessive columns, considering only the ten most frequent categories would be considered. In comparison, the remaining categories would be categorized as "zero" or "not belonging to the ten most frequent categories".

## 3.4. Modeling

At this stage, the prepared data can be analyzed using different data mining methods to achieve the project's main objective and intended outcome. It is necessary to test different methods and compare their outputs. Sometimes it is necessary to return to the previous step and prepare some data algorithms differently to achieve the desired results [52]. In this paper, three different models with three different feature selection methods were considered. Based on prior research on analyzing the relationships between inputs and outputs across different domains, these algorithms have been identified as some of the most widely used and effective methods for identifying variables and associated regression coefficients [58-60]. Feature selection is a popular data preprocessing approach that has shown effectiveness and efficiency in diverse machine learning and data mining applications, as indicated by various studies [61]. Its usage has been observed across multiple domains, including social media [62,63], healthcare [64,65], and biometrics [66,67]. The overall steps of the modeling are as follows:

    I.      Data Preprocessing

- Split the dataset into test, validation, and training sets with a 20%, 20%, and 60% ratio.
- Identify the optimal number of features to include in the algorithm by applying a threshold to the training set to select relevant features.

    II.      Hyperparameter Tuning

- Utilize a grid technique to optimize the algorithm's hyperparameters, considering a predefined set of values for each hyperparameter.
- Consider the validation set to find the most optimal combination of hyperparameters that maximize the model's performance.

    III.      Model Training

- Fit the model to the training set using the optimized hyperparameters obtained from the hyperparameter tuning step.
- To ensure model robustness and generalizability, we utilize cross-validation

    IV.      Model Evaluation

- Employ the trained model to predict the target variable for the test set.
- Evaluate the model's performance by calculating relevant metrics, such as R2 score, mean squared error (MSE), root mean squared error (RMSE), and mean absolute error (MAE), for the training, test, and validation sets.
- Assess the possibility of overfitting or underfitting by comparing the model's performance on different sets.

V. Model Comparison

- Calculate each model's Akaike Information Criterion (AIC) [68] to assess their relative performance and determine the best-performing model.
- Consider the AIC as a critical criterion for model selection, as it accounts for the goodness of fit and model complexity, providing a balanced approach to model comparison.

### 3.4.1. AIC

For better comparison of the best performance of algorithms, the AIC (Akaike Information Criterion) introduced by Akaike in 1973 is used. AIC is calculated using the following formula:

$$AIC = n * ln(RMSE) + 2p \tag{1}$$

Where:

- n is the number of samples (data points).
- p is the number of features used in prediction.

It is important to note that the model with the smallest AIC value is considered the best [68].

### 3.4.2. RMSE

The root-mean-square error (RMSE) measures the deviation between predicted and observed values. It is calculated as a percentage of variance. In academic literature, RMSE is widely used to quantify the accuracy of predictive models. A lower RMSE value indicates a better fit between predicted and observed values [69]. RMSE formula is as follows:

$$RMSE = \sqrt{\frac{1}{N} \sum_{i=1}^{N} (x_i - y_i)^2} \tag{2}$$

Where, $x_i$ is the predicted value and $y_i$ is the observed value [70].

### 3.4.3. Lasso Regression

Lasso is a regularization technique used in regression analysis that performs both subset selection and regression simultaneously. As a result, irrelevant predictor regression coefficients are reduced toward zero, while appropriate variables are selected effectively. This mitigates the overfitting issue often encountered with ordinary least squares regression (OLS) [71].

For each of the $i = 1,2 \ldots, N$, data points, there is a set of multiple predictors $x_{ij} = (x_{i1}, x_{i2}, \ldots, x_{ip})$, where p is the total number of available predictors. OLS is a statistical method for estimating regression coefficients $(\beta_j)$ in a linear regression model [71]. OLS minimizes the mean square error between the predicted values $(\hat{y}_i = \beta_0 + \sum_j \beta_j x_{ij})$ and the observed values $(y_i)$, according to Eq. (3):

$$\hat{\beta} = argmin \left\{ \sum_{i=1}^{N} (y_i - \hat{y}_i)^2 \right\} \quad (3)$$

### 3.4.4. Decision Tree

sThe decision tree algorithm begins with a large dataset. Binary splits are then applied recursively to divide the dataset into smaller and smaller subgroups. At each step, the binary split is defined by one of the independent variables. The decision tree algorithm measures node impurity based on the sum of squared deviations. All possible splits are considered for each independent variable. The split resulting in the smallest sum of squared deviations is chosen. The node impurity measure $(I(d))$ is computed for each node as follows [72]:

$$I(d) = \sum_{i=1}^{N} (y_i - \bar{y}_d)^2 \quad (4)$$

Where, $\bar{y}_d$ represents the sample mean of the dependent variables with respect to the partition. Splitting is determined by the attribute with the least sum of squared deviations. The splitting process continues until there are no more tuples in the dataset or $(I(d))$ is less than a predetermined value.

### 3.4.5. Random Forest

For regression, random forests are formed by growing trees according to a random vector. Therefore, instead of class labels, the tree predictor takes the form of numerical values. As the output values are numerical, the training set is based on a distribution independent of the distribution of the random vector $Y, X$. The mean-squared generalization error for any numerical predictor $h(x)$ is calculated as follows [73]:

$$E_{X,Y}(Y - h(X))^2 \quad (5)$$

### 3.5. Evaluation

Evaluation involves comparing the outcomes with the predetermined business objectives; thus,

interpretation is required to determine the next step. Moreover, a comprehensive review of the entire process should be conducted to identify potential improvement areas [52]. This phase is addressed and discussed in the discussion and result sections of the study.

### 3.6. Stability

An overview of the deployment phase is provided, which may be in the form of a final report or a software component. This phase encompasses activities such as planning the deployment, as well as monitoring and maintenance [52]. This phase is also presented in the discussion and result sections of the study.

## 4. Result

In this section, we began by utilizing data visualization and exploration techniques to gain a better understanding of the data. Next, we employed three feature selection methods, namely Forward Selection, Binary Genetic Algorithm, and Particle Swarm Optimization, to identify the relevant features. We then applied three regression models, namely Lasso, Decision Tree, and Random Forest, to predict the target features. The models' parameters were set using the grid technique, and we compared the nine resulting models using the AIC criterion. To further optimize the Decision Tree model's performance, we used the Genetic Optimization Algorithm to fine-tune its parameters. Finally, we conducted a comprehensive analysis of the decision tree models.

### 4.1. Data Exploration

Through the exploration of the database, valuable information and meaningful relationships can be discovered. Obtaining an overview of the data is especially beneficial during the subsequent modeling process. We conducted data exploration using different strategies, including single-feature, two-feature, and multiple-feature analysis, to uncover hidden and valuable information in the data.

### 4.1.1. Single-feature analysis

Upon analyzing the "number of building units" column, it was observed that over 95% of the projects were single-unit buildings. More than 90% of the projects were categorized as Home Performance projects. Further analysis revealed that approximately 99% of the improvements were focused on the building's body. In comparison, around 1% were related to the ventilation system, and a small percentage were related to the water heater. Notably, 111 buildings were estimated to cost zero dollars after the project. The "Measure Type" column indicated that building improvements were primarily focused on three categories: building body, ventilation system, and water heater, with only six buildings related to water heaters, 667 buildings related to ventilation systems, and the remaining projects related to building bodies.

### 4.1.2. Two-feature analysis

Upon reviewing the data in the columns labeled "Customer Type" and "Total Incentives", it becomes

evident that the average amount of financial incentives provided to owners of government-subsidized buildings is approximately five times higher than the incentives provided to owners of buildings eligible for market-type incentives. This is illustrated in Figure 1.

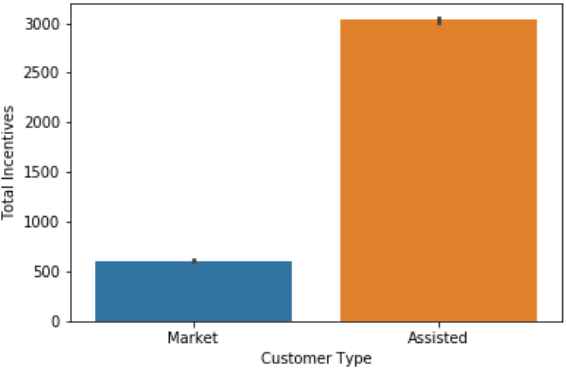

**Figure 1**: The relationship between Customer Type and Total Incentives columns

The analysis of Figure 2 and analyzing the relation between the "Pre-Retrofit Home Heating Fuel Type" factor with target features indicate that buildings that use electricity as their primary heating fuel source have remarkably higher electrical storage than other buildings. The same buildings, however, exhibit negative average values in terms of storage fuel (measured in MMBtu), indicating poor performance in this area. Additionally, buildings that use oil as their primary fuel source incurred the highest savings.

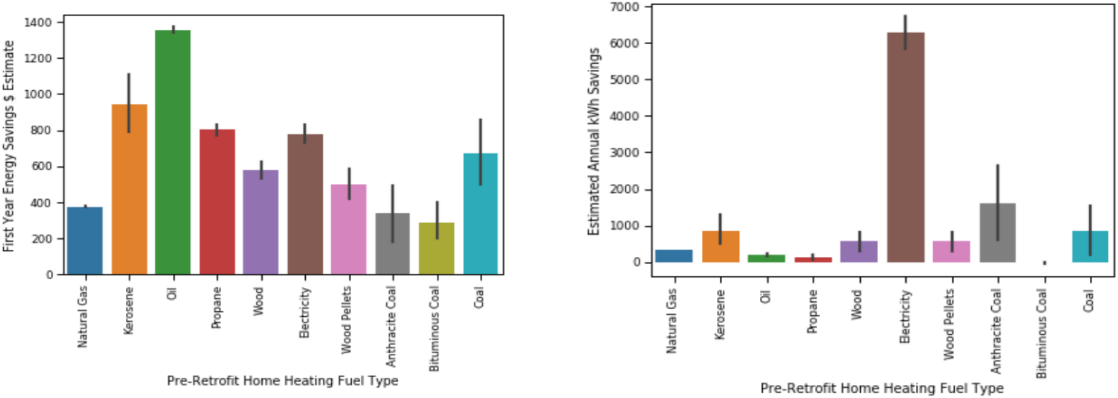

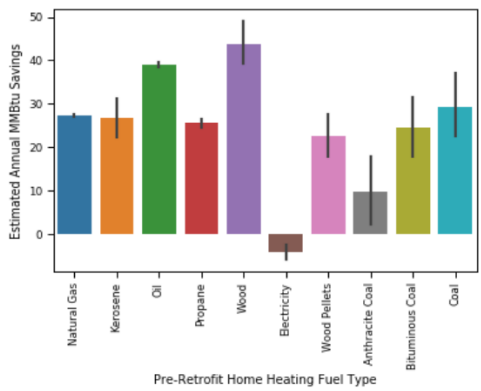 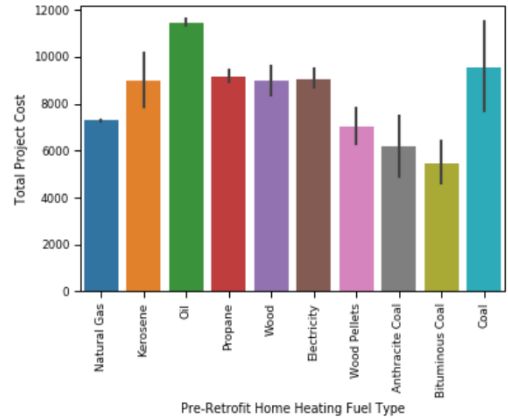

**Figure 2:** Investigation of the "Pre-Retrofit Home Heating Fuel Type" feature

According to Figure 3, after thoroughly analyzing the total project cost concerning the three stated objectives, it becomes apparent that there is a clear and direct correlation between this feature and the targets of fuel storage and cost reduction. However, the same level of correlation was not observed with the reduction of electrical energy consumption, whether in direct or indirect form.

**4.1.3. Multiple-feature analysis**

Figure 4 illustrates the correlation between the features. Fuel storage and cost savings features have a higher correlation (0.64) among the three target variables. Project cost features and loan amount have the strongest correlation (0.65).

Figure 5 illustrates the scatter plot of cost saved and fuel storage along with three features indicating customer type (C_T), whether a Green Jobs Cost Audit plan was received (y/n), and the number of units. The analysis indicates that the 4-unit buildings which received the Green Jobs Cost Audit plan were able to achieve reduced fuel consumption. In general, the 3-unit buildings that received government subsidies (customer type = 1) performed the best in terms of cost savings and reducing fuel consumption.

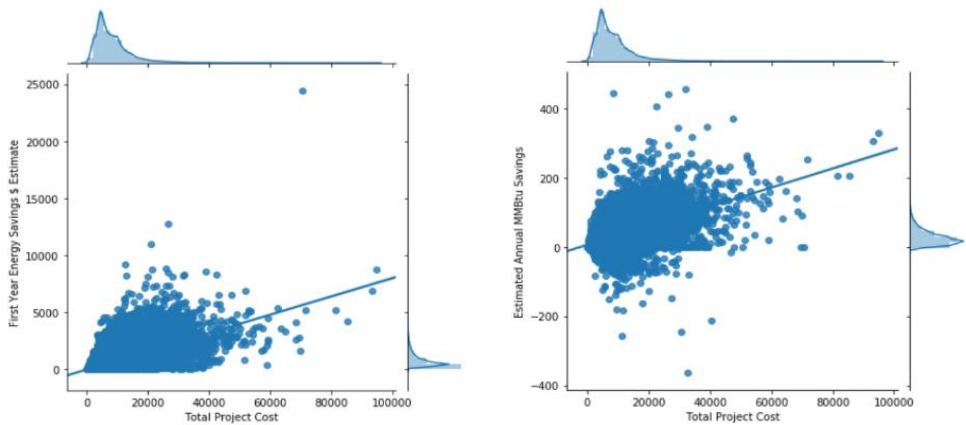

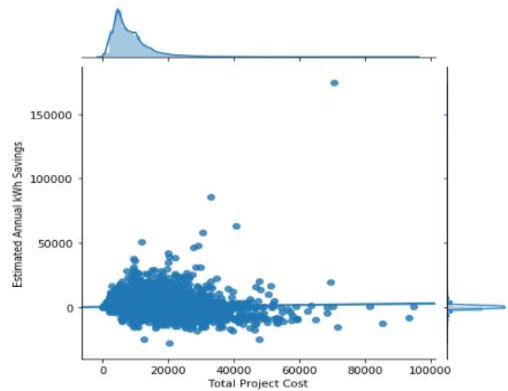

**Figure 3:** The correlation between total project cost and the objectives

## 4.2. Comparison of the Models

We have compared several models for forecasting the three objectives of the problem, utilizing three separate feature selection methods. Tables 3 to 5 provide the Root Mean Square Error (RMSE), the number of selected features, and the AIC criteria for each of the nine models considered. The AIC is applied, which takes into account both the number of features and the level of RMSE. The lower the AIC value, the better the forecasting model. The analysis provides a comprehensive overview of the relative strengths and weaknesses of the different models, enabling us to identify the most effective approach for each of the three objectives.

### 4.2.1. Cost saved

Table 3 demonstrates that in comparison with the other methods, the Forward Selection technique has selected a smaller subset of features. The Genetic Algorithm method resulted in a consistent number of selected features and the lowest RMSE was observed. When combined with Genetic Algorithm and Particle Swarm Optimization for feature selection, the Lasso algorithm produced the best RMSE results. Additionally, the models using Particle Swarm Optimization as the feature selection method showed the most consistent RMSE performance. To further facilitate the comparison of the models based on the AIC criterion, we provide Figure 6, which presents a comprehensive view of the models' relative performance across the feature selection methods.

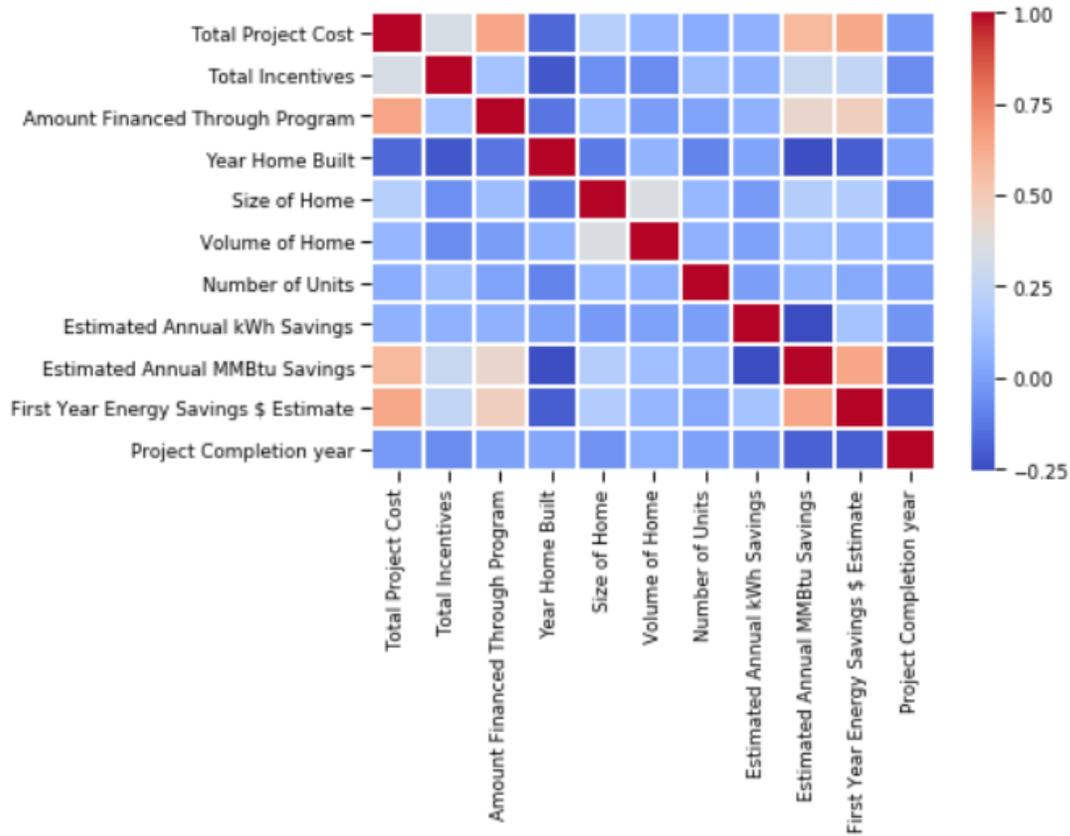

**Figure 4**: The heatmap depicting correlation among features

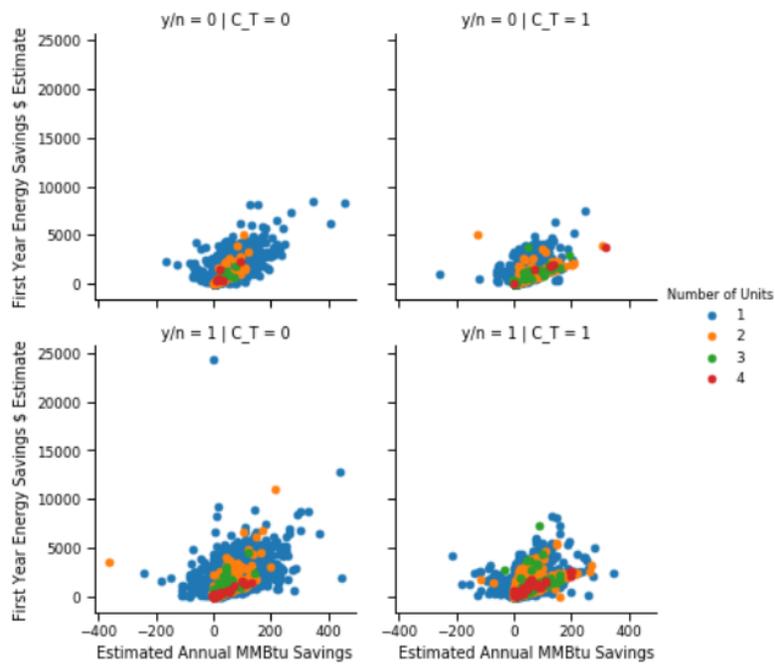

**Figure 5**: The relationship between cost saved and fuel storage, considering customer type, the status of the Green

**Table 3:** Comparison of the models considering different feature selection methods

| Feature selection method | Model | RMSE | Selected Features (%) | AIC |
|---|---|---|---|---|
| Forward feature selection | Lasso | 1505.55 | 8 | 254306.01 |
| Forward feature selection | Decision Tree | 1669.92 | 4 | 257901.13 |
| Forward feature selection | Random Forest | 1086.15 | 11 | 242964.17 |
| Genetic binary algorithm | Lasso | 450.82 | 45 | 212459.47 |
| Genetic binary algorithm | Decision Tree | 563.29 | 44 | 220202.57 |
| Genetic binary algorithm | Random Forest | 460.88 | 45 | 213229.61 |
| Particle swarm optimization | Lasso | 1534.47 | 47 | 255035.37 |
| Particle swarm optimization | Decision Tree | 1587.31 | 47 | 256211.96 |
| Particle swarm optimization | Random Forest | 1596.27 | 31 | 256379.46 |

According to Figure 6, the Lasso and RFs models exhibit pretty similar performance. Considering feature forward selection, the Lasso regression model demonstrates the most favorable performance with the lowest AIC value. This model and the random forest method utilizing particle swarm optimization are the most effective predictive models. On the other hand, the decision tree algorithm does not perform as well as the other two methods.

### 4.2.2. Electrical energy

Table 4 illustrates that compared to the other techniques, the Forward Selection approach has chosen a smaller set of features and achieved the lowest RMSE. Among all three feature selection methods, Random Forest yielded the most satisfactory RMSE results compared to the other algorithms. Furthermore, models utilizing Particle Swarm Optimization as the feature selection method exhibited superior performance in terms of RMSE. Figure 7 provides a comprehensive overview of the relative performance across the methods used for selecting features based on the AIC.

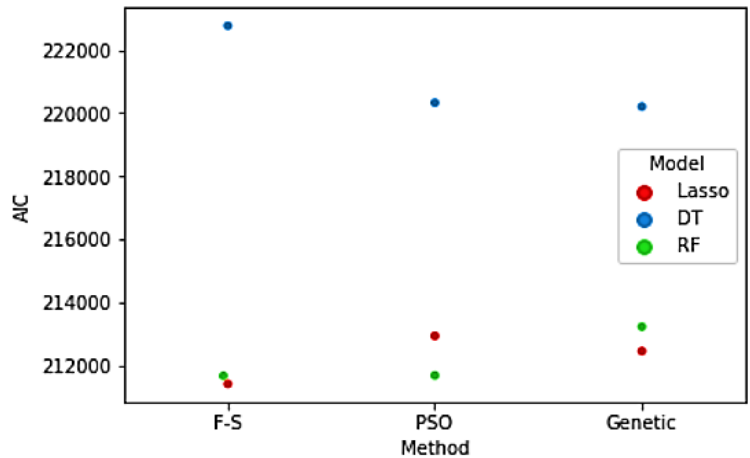

**Figure 6:** Comparison of models based on AIC criteria for predicting the cost savings

**Table 4:** Comparison of the models for electrical energy considering different feature selection methods

| Feature selection method | Model | RMSE | Selected Features (%) | AIC |
|---|---|---|---|---|
| Forward feature selection | Lasso | 438.11 | 16 | 2114118.16 |
| Forward feature selection | Decision Tree | 607.55 | 5 | 222761.49 |
| Forward feature selection | Random Forest | 401.5 | 10 | 211674.05 |
| Genetic binary algorithm | Lasso | 1579.67 | 34 | 256023.16 |
| Genetic binary algorithm | Decision Tree | 1842.40 | 52 | 261399.24 |
| Genetic binary algorithm | Random Forest | 1497.71 | 38 | 254176.50 |
| Particle swarm optimization | Lasso | 457.12 | 52 | 21294.14 |
| Particle swarm optimization | Decision Tree | 565.44 | 44 | 220325.17 |
| Particle swarm optimization | Random Forest | 440.79 | 49 | 211688.31 |

From Figure 7, it can be concluded that the RF model, which utilizes forward selection, performs significantly better than the other models. The Lasso regression model appears to have relatively similar performance across all three feature selection methods. Finally, the decision tree model with genetic algorithm yields the least desirable results with the highest AIC value.

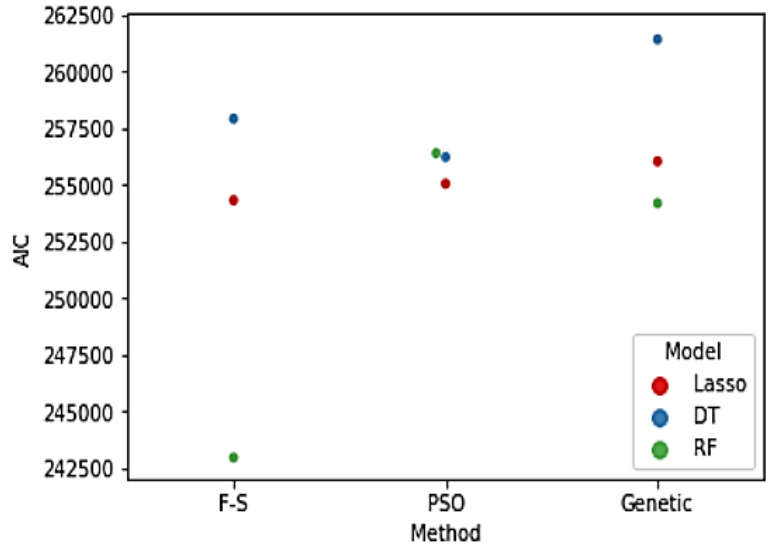

**Figure 7:** Comparison of models based on AIC criteria for predicting the electrical energy consumption

### 4.2.3. Primary Fuel

Table 4 demonstrates that similar to the case with other targets, the Forward Selection technique has opted for a reduced number of features. Among the nine models, the RMSE values are fairly similar, with the best-performing model being the Random Forest (19.87) and the Decision Tree (25.31) performing the worst, which has the fewest features, both utilizing forward selection. Figure 8 offers a comprehensive summary of the comparative performance of the feature selection methods based on AIC.

**Table 5:** Comparison of the models for primary fuel usage considering different feature selection methods

| Feature selection method | Model | RMSE | Selected Features (%) | AIC |
|---|---|---|---|---|
| Forward feature selection | Lasso | 20.22 | 12 | 104511.89 |
| Forward feature selection | Decision Tree | 25.31 | 8 | 112307.11 |
| Forward feature selection | Random Forest | 19.87 | 56 | 103903.04 |
| Genetic binary algorithm | Lasso | 20.35 | 56 | 104811.61 |
| Genetic binary algorithm | Decision Tree | 22.86 | 37 | 108832.72 |
| Genetic binary algorithm | Random Forest | 20.85 | 44 | 105645.56 |
| Particle swarm optimization | Lasso | 20.98 | 45 | 105860.10 |
| Particle swarm optimization | Decision Tree | 23.37 | 45 | 109612.52 |
| Particle swarm optimization | Random Forest | 20.42 | 49 | 104618.76 |

Based on Figure 8, it can be inferred that the Random Forest (RF) model, which uses forward selection, outperforms the other models. The Lasso regression model, which also employs the same feature selection method, ranks second. Random Forest with Particle Swarm Optimization and Lasso with Genetic Algorithm yield comparable results. In contrast, the Decision Tree models demonstrate the poorest performance and have the highest AIC values.

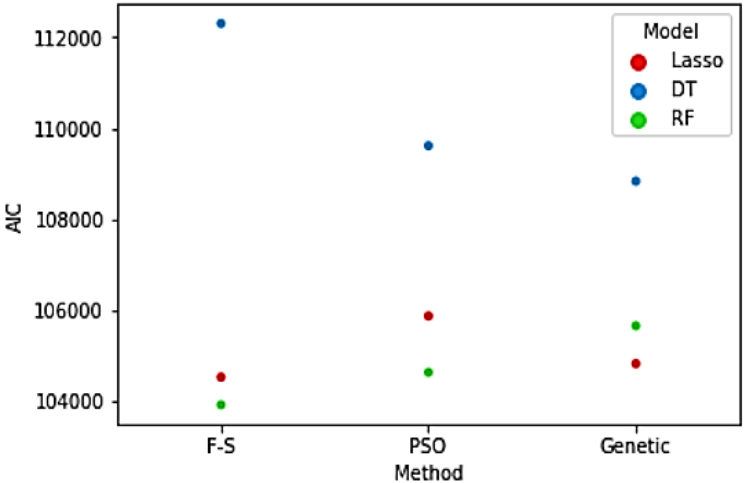

**Figure 8:** Comparison of models based on AIC criteria for predicting the primary fuel usage

## 4.3. Optimization of Decision Tree Algorithm using Genetic Algorithm

We used a genetic algorithm to improve the performance of the decision tree algorithm for prediction, as the original algorithm performed poorly. To execute the GA, we employed the DEAP library [74] and considered a complete list of parameters related to the decision tree, along with their possible values. We began by creating chromosomes and defining separate populations. The Creator function evaluated fitness and defined a single chromosome. Since our populations were not homogeneous, we created custom individuals using the tools function, Initcycle, which determined the possible values of different genes on the chromosome. Rather than initializing the population with attributes, we filled it with individuals, creating a bag full of a specific number of individuals in no particular order. While DEAP has internal functions for mutation and evaluation, we had to define custom functions due to chromosome heterogeneity. Tournament selection involved selecting a user-defined number of chromosomes and running matches between them. The winner of each tournament was the most suitable chromosome, which was then transferred to the crossover. A custom mutation function was called, randomly selecting and mutating one of the individual chromosome genes assigned to it. The modified individual was then returned. The parameters transferred to the decision tree models were evaluated for each individual chromosome, and the resulting MSE score was used as the fitness score. Finally, we combined these functions to create GA, specifying population size, probability of crossing, probability of mutation, and the number of generations. Larger populations allowed for more exploration of the search space but required more computational time. We used the genetic algorithm for each target column to determine the best decision tree, and the findings are presented in the table below.

Table 6: Improvement of the decision tree algorithm with the help of the genetic algorithm

| Feature | AIC(grid) | AIC(Genetic) |
|---|---|---|
| First-Year Energy Savings $ Estimate | 220202.57 | 216336.48 |
| Estimated Annual kWh Savings | 256211.96 | 254295.17 |
| Estimated Annual MMBtu Savings | 108832.72 | 107336.28 |

## 4.4. Decision Tree Analysis

Decision trees, which have their roots in machine learning theory, are effective tools for solving classification and regression problems. A decision tree regression approach is based on the implicit assumption that relationships between features and target objects are either linear or nonlinear [75]. For this analysis, the algorithms with the fewest features have been considered.

### 4.4.1. Cost saved

The decision tree analysis revealed that the average saving cost was $624.67. Saint Lawrence was one such gas provider. Group buildings, which did not use pre-refined fuel, were funded and had Hudson as their gas supplier, which had the lowest reserves (average 316.50). Comparing these two groups, we found that the type of fuel has the greatest impact on storage costs, and the gas supplier is also influential. The overall visualization (up to three level) can be found in Figure 9.

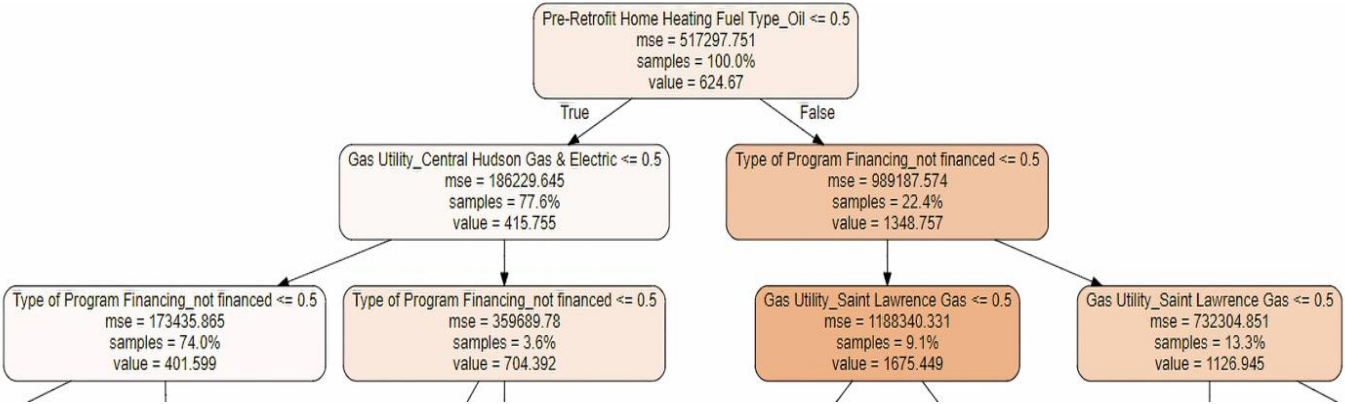

Figure 9: Decision tree visualization plot for the cost saved target

### 4.4.2. Electrical energy

The decision tree analysis, with the overall visualization (up to three level) shown in Figure 10, revealed that the average electrical reserve was 440.18 Kwh. Financial support had been provided. Group buildings, which had their gas suppliers on Long Island but did not receive electricity from them and were also using fuel before the electricity reform, had the lowest electrical reserves (average -8956.6). Comparing these two groups, we found that gas suppliers and financial support significantly impact the number of electrical reserves.

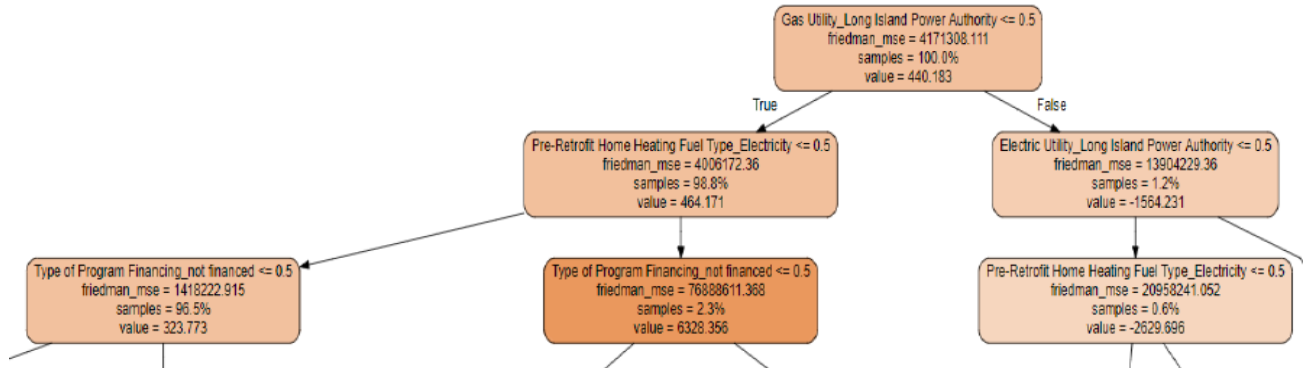

**Figure 10:** Decision tree visualization plot for the electrical energy target

### 4.4.3. Primary fuel usage

Based on the decision tree analysis, the average annual primary fuel reserve was 29.08 MMBTU. There was no electricity reform, the buildings were located in the Jefferson area, and their gas supplier was not Long Island. Comparing the two groups, we found that financial support, the year the project was completed (before or after 2016), and the type of fuel used before the energy reform all played a significant role in storage. Figure 11 provides the overall view of the model up to three levels.

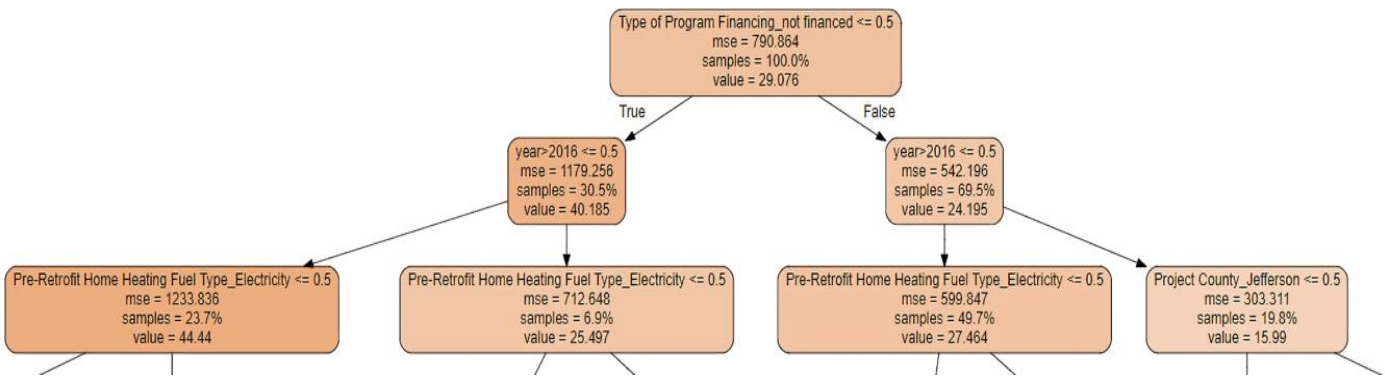

**Figure 11**: Decision tree visualization plot for the primary fuel usage target

## 5. Discussion

Table 7 presents the effective variables of each model for all three target columns. This section considers the most effective of the three feature selection modes as well as the decision tree algorithms after optimization.

Table 7: Influential factors of each model on the target variables

| Target | Cost saved | | | Electrical energy | | | MMBTU | | |
|---|---|---|---|---|---|---|---|---|---|
| Feature/Model | Lasso | DT | RF | Lasso | DT | RF | Lasso | DT | RF |
| Project County | * | * | | | * | * | * | * | |
| Project City | | * | | | * | * | | * | |
| Gas Utility | * | * | | * | * | * | * | * | |
| Electric Utility | | | * | * | * | * | | | |
| Project Completion year | * | | | | | | * | | |
| Customer Type | | | | | * | | | * | |
| Low-Rise or Home Performance Indicator | | | | | * | | | * | |
| Total Project Cost | * | | * | | | | * | * | * |
| Total Incentives | | | * | * | * | | * | | * |
| Type of Program Financing | * | | | * | * | * | | | |
| Amount Financed Through Program | * | * | * | * | * | | * | | * |
| Pre-Retrofit Home Heating Fuel Type | * | * | * | * | * | * | * | * | * |
| Year Home Built | * | * | * | | * | | * | * | * |
| Size of Home | | | * | | * | | | * | * |
| Volume of Home | * | | | | | | * | | * |
| Number of Units | | | | | * | | | * | |
| Measure Type | | | | | | * | | * | |
| Homeowner Received Green Jobs-Green NY Free/Reduced Cost Audit (Y/N) | | | | | | | | | |

According to Table 7, the type of fuel used, financial support, gas and electricity suppliers, total cost of the project, year of construction, and size of the home are found to be the most influential factors. Buildings that rely on electricity as their primary fuel source have not been successful in reducing fuel consumption even after making corrections, so investing in such buildings to reduce energy consumption is not recommended. Financial support plays a crucial role in the success of building energy efficiency projects, as buildings that receive funding have the potential to save more electrical energy and fuel, and therefore costs, compared to those that do not receive financial support. The choice of gas and electricity suppliers is also significant, as buildings with Long Island Power Authority as their electricity supplier do not perform well in electrical energy consumption but are the most suitable option for cost and fuel reduction compared to other suppliers. Furthermore, buildings with higher-than-average total costs have more potential for energy savings and fuel reduction than those with lower costs, which could potentially make them more attractive for loans. Older buildings built before 1980 also have the potential to reduce energy consumption, fuel consumption, and costs saved. Additionally, larger homes with over 540 square feet are generally better at reducing fuel consumption compared to smaller homes. Therefore, it is recommended to invest in larger homes to achieve greater primary fuel consumption reduction.

**6. Conclusion**

In this study, we use three different algorithms with three different feature selection methods to identify the factors that affect energy consumption and costs. The results show that the type of fuel used, financial support, gas and electricity suppliers, total cost of the project, year of construction, and size of the home are all important factors. Specifically, the type of fuel used before modification has the greatest impact on all three target fields.

The analysis showed that different models and different feature selection methods can have a significant impact on the accuracy of the prediction. This is because different models and feature selection methods can extract different information from the data, which can lead to different predictions. This emphasizes the importance of choosing the right model and feature selection method. We also highlight the importance of using an efficient hyperparameter selection method and how it can improve the performance of an algorithm.

It should be noted that this study has certain limitations. Data used in the study lacks information regarding consumption culture, climate, level of education of residents, the age range of residents, occupancy status, and initiatives to reduce energy consumption and associated costs. Future studies may consider employing alternative data mining techniques and gathering additional data to address these limitations.